# Lexical Normalisation of Twitter Data

Bilal Ahmed
Department of Computing and Information Systems
The University of Melbourne
Victoria, Australia
Email: bahmad@student.unimelb.edu.au

*Abstract*—Twitter with over 500 million users globally, generates over 100,000 tweets per minute[1]. The 140 character limit per tweet, perhaps unintentionally, encourages users to use shorthand notations and to strip spellings to their bare minimum "syllables" or elisions e.g. "srsly". The analysis of Twitter messages which typically contain misspellings, elisions, and grammatical errors, poses a challenge to established Natural Language Processing (NLP) tools which are generally designed with the assumption that the data conforms to the basic grammatical structure commonly used in English language. In order to make sense of Twitter messages it is necessary to first transform them into a canonical form, consistent with the dictionary or grammar. This process, performed at the level of individual tokens ("words"), is called lexical normalisation. This paper investigates various techniques for lexical normalisation of Twitter data and presents the findings as the techniques are applied to process raw data from Twitter.

I. INTRODUCTION

A Twitter message or "tweet" consists of 140 or fewer characters, and generally contains hash tags and @ symbols. In order to lexically analyse a Twitter message each token needs to be identified on a case by case basis before normalisation techniques are applied to correct spelling mistakes and make sense of the various acronyms and elisions frequently used in Twitter messages. The proceeding sections describe in detail the various techniques that are applied to identity: "known" or "in-vocabulary" words; punctuation and special symbols (both general and Twitter specific); and candidates for normalisation. We then apply various normalisation techniques to correct out of vocabulary ("OOV") tokens.

II. IN VOCABULARY TOKENS

The first step is to identify tokens or words that are in vocabulary. The dictionary is searched for an exact match of the word. A token is tagged as "in vocabulary ("IV") if an exact match is found. For the purpose of this project we have used a lexicon of 115,326 words (words.txt) to identify "in vocabulary" words. Tokens that fall outside of this vocabulary are then considered as candidates for normalisation and are further processed or marked as non-candidates if deemed not fit for normalisation.

III. NON-CANDIDATE TOKENS

In addition to common punctuation symbols a Twitter message or "tweet" generally contains hash tags, the "#" symbol, to mark keywords or topics in a tweet and the "@" symbol followed by a user's Twitter username to refer to a user when replying or commenting. The tokens are parsed using regular expression to identify special characters, punctuation and Twitter specific symbols. These special tokens are marked as non-candidates ("NO") and are not processed for normalisation.

IV. NORMALISATION OF OUT OF VOCABULARY TOKENS

Lexical normalisation is the process of transforming tokens into a canonical form consistent with the dictionary and grammar. These tokens include words that are misspelt or intentionally shortened (elisions) due to character limit in case of Twitter.

When a word falls outside the vocabulary as defined by the collection of words in *word.txt* file, and does not contain any special characters, punctuation or Twitter specific symbols, it is marked as out of vocabulary ("OOV") and is processed as a candidate for normalisation.

**Overview of the Normalisation Process:** Once a candidate has been identified for normalisation, **firstly**, edit distance (Levenshtein distance) technique is applied to find matches from (words.utf-8.txt) which are within 2 (inclusive) edit distance of the query. The results are stored in an array. We refer to this set as the "First Set of Matches based on Edit Distance" since they contain approximate matches based on their textual similarity to the query.

**The second step** in the process is to apply Refined Soundex technique to this set of matches based on edit distance. This refines the set and results in approximate matches that are phonetically similar to the query. The results are stored in another array. This refined and phonetically similar set of words is referred to as "Phonetic Matches".

---

[1] Twitter in numbers, The Telegraph, March 2013

**The third step** is to find an approximate match using Peter Norvig's Algorithm. This returns one match deemed closest to the query by the algorithm.

**The forth step** compares the result of the Peter Norwig algorithm with those obtained in Step 2 by applying Refined Soundex technique. If both of the results are same, i.e. only 1 phonetic match is found by Refined Soundex technique and is the same as that returned by Peter Norvig's Algorithm, then no further processing is performed, and the result is used as the normalised version for the query. If more than 1 phonetic match is returned by Refined Soundex technique then based on rules described in Section 4.3 a further 5-Gram Context Matching is performed.

**The fifth step** is to perform a 5-Gram Context Matching technique using **each phonetic match** as the query in the following regular expression:

{Previous word, **Query**, Next word}

This takes into account the previous and next words, to the query, and performs an exhaustive search to find the most commonly pattern. This technique **uses each phonetic match** to see if it is a likely candidate based on its occurrence as defined by the pattern above. This technique is further explained in Section 4.4. The outcome of this search is used as the normalised version for the query.

TABLE 1: CORPORA UTILIZED FOR NATURALISATION OF OUT OF VOCABULARY "OOV" WORDS.

| Corpus[2] | Features |
|---|---|
| words.txt | 115,326 words[*] |
| words.utf-8.txt | 645,288 words[+] |
| big.txt | 1 M words[^] |
| w5_.txt | 1 M words[#] (1,044,268) |

Approximate matching techniques are performed to extract relevant matches from over 3.5 M words contained in corpora listed in Table 1.

The following sections explain in detail the techniques utilised to find the closest matches to "out of vocabulary" (OOV) tokens.

*A. Edit Distance (Section 4.1, Step 1)*

The OOV or "query" token is compared against the 645,288 words contained in the words.utf-8.txt file. The first set of crude matches is gathered by calculating the *Levenshtein distance* between the query and the words in the dictionary.

**Levenshtein distance** is a string metric for measuring the difference between two sequences. It is defined as the minimum number of single character edits (insertion, deletion, substitution) required to change one word into the other. The phrase edit distance is often used to refer specifically to Levenshtein distance.

Using this technique which employs matching based on the textual representation of the characters in the query; the dictionary (words.utf-8.txt) is searched for matches that are within 2 (inclusive) Levenshtein distance of the query. This generates the first set of approximate matches based on the textual similarity to the query. The results are stored in an array. This set generally contains words that may have been misspelt in the query.

*B. Phonetic Matching (Section 4.2, Step 2)*

Phonetic matching algorithms match two different words with similar pronunciation to the same code. These algorithms compare and index words that are phonetically similar and can be used for spelling correction.

**Refined Soundex** algorithm is an improvement to the original Soundex algorithm, in which the letters are divided into more groups (Fig 1.) based on their sound. Also, the length of the result is not truncated, so the code does not have a fixed length. This provides better resolution for phonetic matching as compared to the original Soundex.

Orginal Soundex

| b, f, p, v | 1 |
| c, g, j, k, q, s, x, z | 2 |
| d, t | 3 |
| l | 4 |
| m, n | 5 |
| r | 6 |

Refined Soundex

| b, p | 1 |
| f, v | 2 |
| c, k, s | 3 |
| g, j | 4 |
| q, x, z | 5 |
| d, t | 6 |
| l | 7 |
| m, n | 8 |
| r | 9 |

Fig. 1. Phonetic Matching

Refined Soundex is used to further analyse and phonetically match words gathered in the first set, based on their Levenshtein distance to the query as described in Section 4.1. The words in the array are filtered based on their phonetic similarity to the query as shown in Figure 2 below.

---

[2] [*]words.txt is used to search for in-vocabulary words.
[+]words.utf-8.txt is used to search for approximate matches for normalisation of "OOV" words
[^]big.txt consists of about a million words. The file is a concatenation of several public domain books from Project Gutenberg and lists of most frequent words from Wiktionary and the British National Corpus. This is used by Peter Norvig's algorithm for naturalisation.
[#]w5_.txt is used to perform context based 5-Gram matching

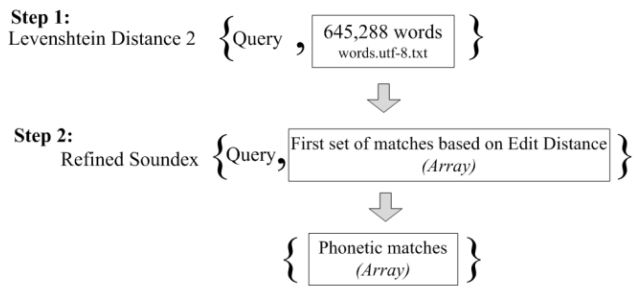

Fig. 2. Steps 1 and 2

This produces an array containing closer matches and is a set of words which are:
i. Phonetically similar and,
ii. Within 2 or less Levenshtein distance to the query

*C. Peter Norvig's Algorithm (Section 4.3, Step 3)*

**Peter Norvig's** Algorithm generates all possible terms with an edit distance of less than or equal to 2 (which includes deletes, transposes, replaces, and inserts) from the query term and searches them in the dictionary (big.txt, see Table1).

For a word of length n, an alphabet size a, an edit distance d=1, there will be n deletions, n-1 transpositions, a*n alterations, and a*(n+1) insertions, for a total of 2n+2an+a-1 terms at search time. This is much better than the naive approach, but still expensive at search time (114,324 terms for n=9, a=36, d=2) and is language dependent. Because the alphabets are used to generate the terms, and are different in many languages, it could potentially lead to a very large number of search terms. E.g. In Chinese: a=70,000 Unicode Han characters. Never the less, it usually achieves 80-90% accuracy averaging at about 10 words per second.

For the purpose of this experiment we apply Peter Norvig Algorithm to find the best match for a given query as shown in Figure 3, below.

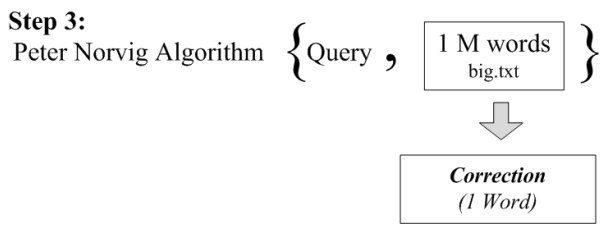

Fig.3. Step 3 – Peter Norvig Algorithm

The result is then compared (Figure 4) with the phonetically matched words derived in Section 4.2 based on the following rules:

a) Peter Norwig's result takes precedence and is returned as the normalised word for a query, if 0 phonetic matches are found after applying Refined Soundex algorithm (in Section 4.2).

b) If both Refined Soundex and Peter Norwig algorithm derive the same result, i.e. only 1 phonetic match is found which, is the same as Peter Norwig's result, then no further processing is conduced and the result is returned as the normalised version for a query.

c) If Refined Soundex returns more than 1 phonetic match, then the query is further analysed using 5-Gram Context Matching technique as detailed in Section 4.4.

**Step 4:**
Comare Results from Steps 2 and 3

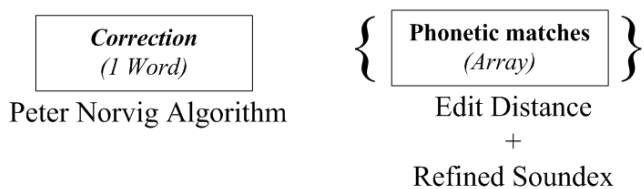

Fig. 4. Step 4

*D. 5-Gram Context Matching (Section 4.4, Step 5)*

If there are more than 1 phonetic matches found, in other words if Refined Soundex technique (Section 4.2) returns more than one phonetic match then a 5-Gram Context Matching technique is applied using **each phonetic match** as the query in the following regular expression:

{Previous word,  **Query**, Next word}

The following rules are applied to assemble the regular expression for 5-Gram Matching:

a) If the previous and next words to the query are both in vocabulary, then following pattern is used: {Previous word,  Query, Next word}

b) If only the previous word is in vocabulary then: {Previous word,  Query} is used.

c) Else if only the next word is in vocabulary then: {Query, Next word} is used

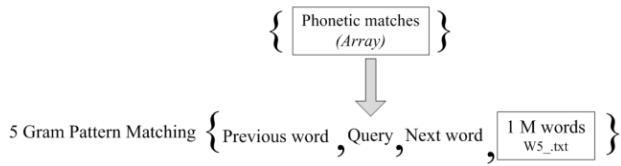

Fig.5. Step 5 – 5-Gram Context Matching

The query which returns the maximum number of occurrences in w5_.txt (which consists of over a million words as 5-Grams) is returned as the normalised version of the query as shown in Figure 5. Here the most common occurrence of the {Previous word, **Query**, Next word} is returned as the result, where each phonetic match is used as query to find a likely candidate based on its occurrence as defined by the pattern above

## V. CONCLUSION

Normalising tokens with high accuracy can be quite a challenge given the number of possible variations for a given token. This is further compounded by the ever increasing and evolving elisions and acronyms frequently used in social media tools such as Twitter. It is important to take into consideration the various normalisation techniques that are available and to pick the ones that best suit the purpose. A blend of techniques such as edit distance and Soundex or Refined Soundex usually results in better accuracy as compared to their standalone application. Techniques based on context such as Peter Norvig's algorithm increase the accuracy of normalisation. Similarly, N-Gram matching, although exhaustive, can be optimised to produce accurate results based on the context.

### ACKNOWLEDGMENT

The author would like to thank Professor Rao Kotagiri and Jeremy Nicholson for their invaluable guidance throughout this research. The author would also like to thank Professor Justin Zobel.